\pdfoutput=1

\documentclass[11pt]{article}

\usepackage[preprint]{acl}

\usepackage{times}

\usepackage{array}
\usepackage{bm}
\usepackage{latexsym}
\usepackage{mathrsfs}
\usepackage{amsfonts}
\usepackage{amssymb}
\usepackage{amsmath}
\usepackage{multirow}
\usepackage{bbding}
\usepackage{booktabs}
\usepackage{tabularx}
\usepackage{color}
\usepackage{graphicx}
\usepackage{float}

\newcolumntype{C}[1]{>{\centering\arraybackslash}m{#1}}

\usepackage[T1]{fontenc}
\usepackage[utf8]{inputenc}
\usepackage{microtype}

\title{AIMA at SemEval-2024 Task 10: \\History-Based Emotion Recognition in Hindi-English Code-Mixed Conversations}

\author{\begin{tabular}{c}
Mohammad Mahdi Abootorabi$^{\diamond}$, Nona Ghazizadeh$^{\diamond}$, Seyed Arshan Dalili$^{\diamond}$,\\ Alireza Ghahramani Kure$^{\diamond}$, Mahshid Dehghani$^{\diamond}$, Ehsaneddin Asgari$^{\mathsection}$
\end{tabular}\\$^{\diamond}$ NLP \& DH Lab, Computer Engineering Department, Sharif University of Technology \\ $^{\mathsection}$ Qatar Computing Research Institute, Doha, Qatar\\
\texttt{\{mahdi.abootorabi, nona.ghazizadeh, seyedarshan.dalili,}
\\
\texttt{a.ghahramani, mahshid.dehghani\}@sharif.edu}
\\
\texttt{easgari@hbku.edu.qa}
}

\begin{document}
\pagenumbering{gobble}
\maketitle
\begin{abstract}
In this study, we introduce a solution to the SemEval 2024 Task 10 on subtask 1, dedicated to Emotion Recognition in Conversation (ERC) in code-mixed Hindi-English conversations. ERC in code-mixed conversations presents unique challenges, as existing models are typically trained on monolingual datasets and may not perform well on code-mixed data. To address this, we propose a series of models that incorporate both the previous and future context of the current utterance, as well as the sequential information of the conversation. To facilitate the processing of code-mixed data, we developed a Hinglish-to-English translation pipeline to translate the code-mixed conversations into English. We designed four different base models, each utilizing powerful pre-trained encoders to extract features from the input but with varying architectures. By ensembling all of these models, we developed a final model that outperforms all other baselines.
\end{abstract}

\section{Introduction}
The first subtask of SemEval 2024 Task 10 focuses on Emotion Recognition in Conversation (ERC)
\cite{kumar-etal-2023-multilingual}. This subtask requires the design of a model capable of predicting an emotion for each utterance. 
Our final system is an ensemble of four high-performing models we developed in this paper.
Our primary strategy involves leveraging powerful pre-trained models and utilizing
the context of preceding and succeeding utterances in the conversation. We also consider the sequential information of the conversation to accurately predict emotions. The final system is designed to work with Hindi-English code-mixed conversations. A detailed description of the task is available in \citep{Kumar2024SemEval}.

ERC is an emerging research frontier in Natural Language Processing (NLP), that aims to identify emotions in conversational data. The ability to accurately recognize emotions in conversation is crucial for a variety of applications, including opinion mining from social media platforms \citep{poria2019emotion}.
ERC is also extremely important for generating emotion-aware dialogues that require an understanding of the user’s emotions. It is useful in various sectors, such as healthcare for psychological analysis and education to aid in understanding student frustration \citep{Antony2021EmotionRM}.

ERC presents several research challenges due to the complexity and rapid changeability of emotions in conversation. The same words can convey different emotions depending on the context, adding a layer of complexity to the task \citep{kumar-etal-2023-multilingual}. This complexity is further amplified in code-mixed conversations, a common phenomenon in multilingual societies and online social media platforms where two or more languages are used interchangeably. The challenges in ERC for code-mixed conversations include \textbf{(i) Linguistic Complexity}, due to complex linguistic structures and sentence or word-level language switches; \textbf{(ii) Insufficient Training Data}, as the scarcity of annotated datasets hampers the training of deep learning models; \textbf{(iii) Cultural Nuances}, since emotions can be expressed differently across cultures and languages; and \textbf{(iv) Ambiguity and Context-Dependence}, as word meaning and emotions vary based on context and language.

\section{Background}
The official dataset for this task is the MaSaC dataset \citep{masac}, a mixed Hindi-English language dataset relevant to our study of emotion recognition in code-mixed dialogues
\citep{kumar-etal-2023-multilingual}. This dataset consists of approximately $8506$ train, $1354$ validation, and $1580$ test sentences. ERC has garnered significant attention in the NLP community due to its potential applications.

Recent research in ERC has attempted to address these challenges, but there are still many areas for improvement. For instance, most current approaches to ERC focus on text-based data, overlooking the rich emotional information that can be gleaned from other modalities such as voice tone and facial expressions  \citep{kumar-etal-2023-multilingual}.

With recent advances in Large Language Models (LLMs), many works leverage the power of these large models for ERC task \citep{tu-etal-2023-empirical}. \citep{lei2023instructerc} introduced a novel approach, which leverages LLMs to reformulate the ERC task from a discriminative framework to a generative one. This approach has shown significant improvements over previous models on several ERC datasets.
While considerable research has focused on discerning the emotions of individual speakers in monolingual dialogues, understanding the emotional dynamics in code-mixed conversations has received relatively less attention. This is the gap our study aims to address. \citep{kumar-etal-2023-multilingual} proposed an innovative approach that integrates commonsense information with dialogue context to interpret emotions more accurately in code-mixed dialogues. They developed a pipeline based on a knowledge graph to extract relevant commonsense facts and fuse them with the dialogue representation. \citep{wadhawan-aggarwal-2021-towards}, the closest work to ours, introduced a new Hinglish dataset for emotion detection in Hindi-English code-mixed tweets. They trained various deep learning approaches, including transformer-based models, for emotion recognition task performance on this dataset.

This paper aims to delve deeper into the current state of ERC and propose models to take a step toward solving these challenges. Our method differs from existing approaches in that it incorporates both context and sequential information to improve emotion prediction performance.

\section{System Overview}

\subsection{Preprocessing Data}
In the preprocessing stage, we implement a two-step translation process due to the unique nature of our data, which comprises Hindi-English mixed conversations. At present, there are no robust models trained specifically in this mixed language, nor are there translators capable of directly translating Hindi-English mixed text to English with acceptable performance. As a result, we first need to translate our data to English. In the first step, we transform our Hindi-English mixed conversations into Hindi using the indic-trans transliteration module \citep{Bhat:2014:ISS:2824864.2824872}, a tool proficient in cross-transliteration among all Indian languages. Following this, we employ SeamlessM4T Medium \citep{communication2023seamlessm4t} to translate these Hindi conversations into English. The English conversations obtained from this process serve as our preprocessed data.
\subsection{Model Architecture}
In this section, we propose the model architectures that were used to construct our final ensemble model. We designed three distinct architectures, and the final model is an ensemble of four models trained based on these architectures. The second model follows the same architecture as the first, but it is trained on an augmented dataset. Our system predicts the emotion of the current sentence using majority voting based on the predicted emotions from four base models.

\begin{figure*}
\begin{center}
  \includegraphics[width=2\columnwidth]{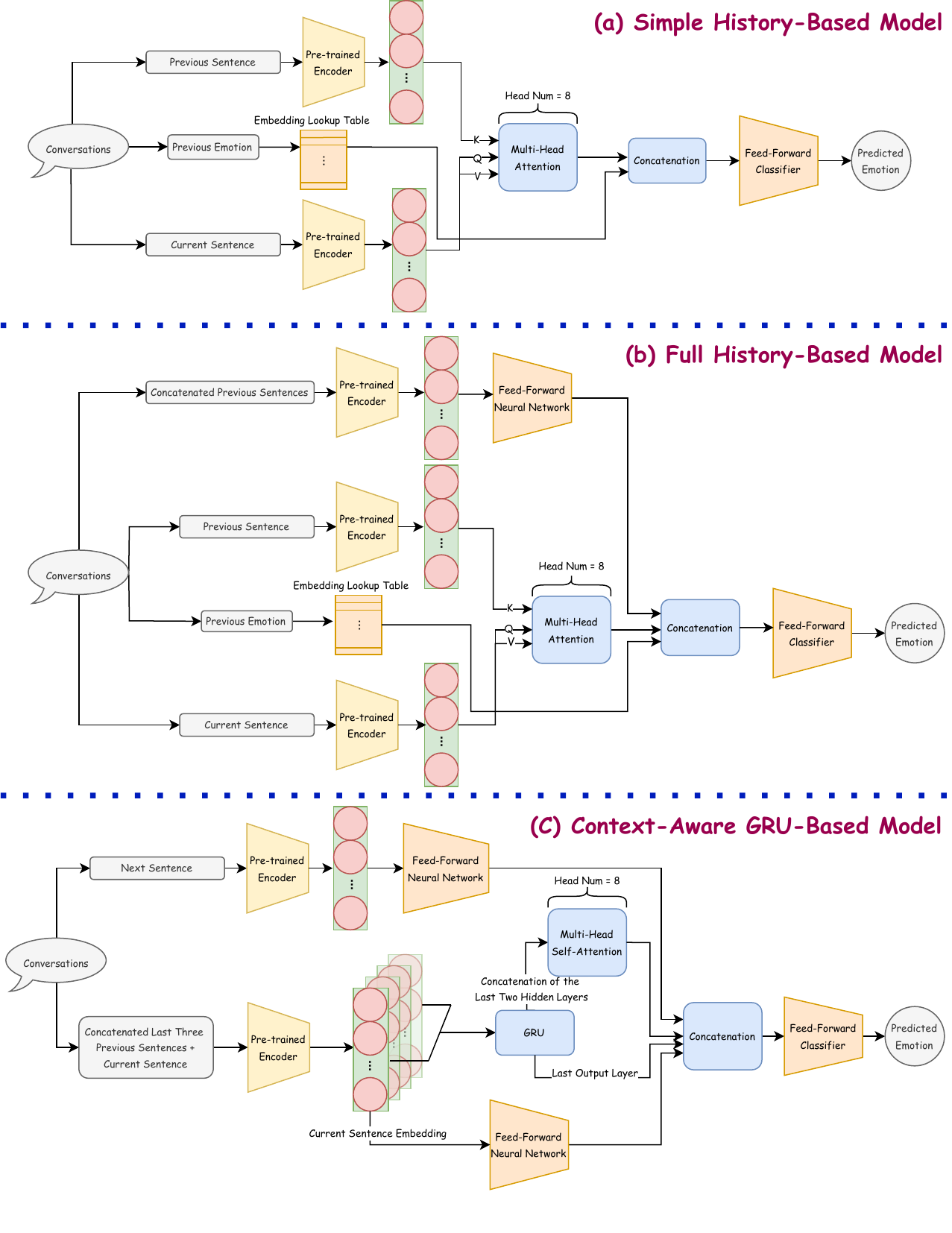}
\end{center}
\caption{
Three proposed base model architectures for predicting the emotion of the current sentence.
(a): This model utilizes only the basic historical information from the conversation. (b): This model leverages information from all past sentences, in addition to the information used in the previous architecture. (c): This model employs GRU to leverage sequential information and incorporates future information to gain a more comprehensive understanding of the context of the current sentence.
}
\label{fig:models}
\end{figure*}

Given the specific domain of the task and the limited number of samples in the dataset, it is crucial to strike a balance between model complexity and performance. Overly complex models may lead to overfitting, especially given the unique distribution of our dataset. Conversely, overly simple models may not capture the complexity of this particular task. Therefore, we aimed to find a balance, ensuring adequate model complexity to learn effectively from the data without leading to overfitting.
Furthermore, due to the limited dataset for this task and the special domain and emotions that are used, such as contempt, we leveraged the encoder component of a pre-trained RoBERTa-based model \citep{liu2019roberta} for the emotion recognition task in sentences, and fine-tuned it for our specific task and domain. This model was trained on the GoEmotions dataset \citep{demszky2020goemotions}, allowing us to employ the capabilities of pre-trained models for our task. This encoder was incorporated into all of our base models for sentence encoding. In the following parts, each of our base models is explained in detail. An overview of architectures is shown in Figure \ref{fig:models}.
\subsubsection{Simple History-Based Model}
This model leverages both the current sentence, for which we aim to predict the emotion, and the preceding sentence along with its associated emotion as historical information to enhance the model’s prediction. Both the current and previous sentences are processed through our pre-trained encoder to obtain their respective embeddings.
We then employed a multi-head attention mechanism
\citep{NIPS2017_3f5ee243}
with 8 heads. In this mechanism, the keys are derived from the embeddings of the previous sentence, while the queries and values are derived from the current sentence. The use of 8 heads in the attention mechanism enables the model to capture information from different representational spaces at various levels of abstraction. This design allows the model to focus on the most relevant parts of the current sentence based on the context provided by the previous sentence.

For emotion representation, we utilized a 50-dimensional embedding space learned by our model. The embedding of the previous emotion and the output of the attention mechanism are concatenated and passed through a feed-forward classifier to predict the emotion. This classifier consists of two linear layers, a LeakyReLU activation function, and a Softmax layer for output normalization.

\subsubsection{Simple History-Based Model + Data Augmentation}
This model architecture is identical to the base model described earlier. The key difference lies in the training data. We used a Pegasus paraphrase model \citep{zhang2019pegasus} to augment our dataset and increase its size. We expanded our dataset by randomly selecting three sentences from the first ten paraphrases of each original sentence. Given the limited size of the original dataset, this augmentation method should enhance the model’s learning capability by exposing it to a wider range of data. 

\subsubsection{Full History-Based Model}
This model, which is an extension of the Simple History-Based model, aims to leverage more historical information for improved performance. In addition to the current sentence, previous sentence, and previous emotion, we also incorporated the concatenated string of all previous sentences in the conversation into our model. The rationale behind this is to enable the model to access additional information and gain a better understanding of the context of the current sentence within the conversation.
The concatenated string of all previous sentences is processed through our pre-trained encoder to obtain the history embedding. This encoding is then passed through a simple feed-forward neural network, which consists of two linear layers, a batch normalization layer, a dropout layer, and a LeakyReLU activation function. This network transforms the 768-dimensional input into a 128-dimensional space.

The processing for the current sentence, previous sentence, and previous emotion remains the same as in the Simple History-Based Model. For the classifier network, we concatenated the output of the feed-forward network for previous sentences with the output of the attention mechanism and the emotion embedding. This concatenated vector is then passed to the classifier to predict the current emotion. The classifier comprises three linear layers, a batch normalization layer, two dropout layers, a LeakyReLU activation function, a ReLU activation function, and a Softmax layer for output normalization.

\begin{table*}
\centering
\begin{tabular}{C{7.8cm}|C{1.55cm}C{1.4cm}C{1.55cm}C{1.55cm}}
\hline
\noalign{\vskip 1mm}
Model Name & Weighted F1 & Accuracy & Weighted Precision & Weighted Recall\\
\noalign{\vskip 1mm}
\hline
\noalign{\vskip 1mm}
GPT-3.5 Turbo & 0.2662 & 0.3070 &0.2582 &0.3070\\ 
Decision Tree & 0.2895 & 0.2937 &0.2900 & 0.2937\\ 
Linear Regression & 0.3394 & 0.4405 &0.4117 &0.4405\\ 
Fine-tuned Sentence Emotion Recognition & 0.3683 & 0.4506 &0.3667&0.4506\\ 
\noalign{\vskip 1mm}
\hline
\noalign{\vskip 1mm}
Simple History-Based Model & 0.4018 & 0.4380 &0.4043&0.4380\\ 
Simple History-Based Model + Data Augmentation & 0.3780 & 0.4253 & 0.3712&0.4253\\ 
Full History-Based Model & 0.3992 & 0.4285 & 0.3963&0.4285\\ 
Context-Aware GRU-Based Model &0.4058& 0.4373 & 0.4024&0.4373\\ 
\noalign{\vskip 1mm}
\hline
\noalign{\vskip 1mm}
Final Model (Ensemble) & \textbf{0.4080} & \textbf{0.4430} &\textbf{0.4090}&\textbf{0.4430}\\ 
\end{tabular}
\caption{ Performance of various models on the test dataset. The first group of models represents the baselines. The second group consists of models based on our proposed architectures. The final model, an ensemble of four proposed models, represents the performance of our final system.}
\label{tab:accents}
\end{table*}

\subsubsection{Context-Aware GRU-Based Model}
This model, more complex than its predecessors, introduces several key modifications. Firstly, it incorporates information from both the preceding and succeeding sentences in a conversation, allowing the model to leverage both past and future contexts. Secondly, in contrast to previous architectures that use the emotion of the previous sentence, this model omits this feature to prevent error propagation during the inference phase. If a model incorrectly predicts the emotion of one sentence, it could potentially use this incorrect information when predicting the emotion of the next sentence, leading to further errors. Lastly, this model employs a Gated Recurrent Unit (GRU) \citep{chung2014empirical, cho2014learning}, enabling it to leverage the sequential information in the conversation.

The model processes all sentences up to and including the current one (for which we want to predict the emotion) and the next sentence through our pre-trained encoder to obtain their embeddings. If the current sentence in the conversation has more than three previous sentences, only the last three are considered, making the model focus on the most recent context. These embeddings are then passed through a stacked GRU, consisting of two GRUs with a hidden dimension of 256 and a dropout rate of 0.25.
Both the current and next sentences went through a transformation via a linear layer and a dropout layer to generate output encodings in a common 256-dimensional space. The last two hidden layers of the GRU are concatenated and passed through a multi-head self-attention mechanism, similar to our previous models. 

The output of the last layer of the GRU, the output of the attention mechanism, and the transformed outputs of the current and next sentences are concatenated and passed to a feed-forward classifier to predict the current emotion. The classifier comprises two linear layers, a dropout layer, a LeakyReLU activation function, and a Softmax layer for output normalization.

\section{Experimental Setup}
We utilized the official dataset provided for the task as the only data source for our system. The default split provided for the task was also used. During the development phase, the validation set was exclusively used for evaluating various steps and experimental configurations. For the final submission, models were fine-tuned on both the training and validation splits.
For evaluation purposes, our primary metric was Weighted F1. However, to provide a more comprehensive analysis, we also reported three additional metrics, as detailed in Table \ref{tab:accents}.
Our training process primarily employed the PyTorch and Transformers libraries. All base models were trained using the early stopping method and the AdamW 
\citep{loshchilov2019decoupled}
optimizer. A learning rate scheduler was used, with a lower learning rate set for the pre-trained encoder (5e-6) compared to other parameters (1e-4). The batch size was set to 1 for the Context-Aware GRU-Based model and 4 for other models during training. The cross-entropy loss function was used for the training.

\section{Results}
Table \ref{tab:accents} presents SubTask 1 results. We compare our approach with four baseline models. The first baseline is GPT 3.5 Turbo, for which we used its API key to input the entire conversation and predict the emotion for each sentence. The results of this baseline model illustrate that this task is much more challenging than general sentence emotion recognition because it is domain-specific. The next two models are traditional ones, namely Linear Regression and Decision Tree, that utilize embeddings extracted from the LaBSE sentence encoder \citep{feng-etal-2022-language}. The LaBSE model serves as a powerful encoder for our text data, enabling us to achieve comprehensive and multilingual text embeddings. The final baseline model is similar to a Simple History-Based model. It employs our pre-trained encoder but does not use any context, such as the previous sentence, and relies solely on the current sentence.

Moving on to the comparison of our models, we first consider the Simple History-Based model. By comparing its results with the Full History-Based model, we find that most of the information for predicting the emotion is contained in the current and the previous sentence. Therefore, information from all of the previous sentences is not as useful for predicting emotion. Our second model, which uses data augmentation, does not perform well. This is likely due to overfitting and the domain-specific nature of the conversations, making data augmentation less effective. As can be seen in our models, the Context-Aware GRU-Based model outperforms the others. This is because it incorporates information from both the preceding and succeeding sentences and the GRU can leverage the sequential information in the conversation. The closeness of the results between the Context-Aware GRU-Based model and the Simple History-Based model reinforces our assumption that most information for predicting emotion is in the current and previous sentence. All of our models outperform the baselines. For our final model, we create an ensemble of these four models using majority voting. This ensemble model outperforms each individual model, achieving an F1-score of 0.4080.

\section{Conclusion}
In this paper, we proposed a novel method to address the Code-Mixed Emotion Recognition in Conversations (ERC) challenge. Our approach leverages the power of pre-trained large models and incorporates both previous and future context information of the current utterance, as well as sequential information of the conversation up to that point, to recognize each utterance’s emotion. In addition to our primary model, we utilized other base models with different architectures based on various Deep Learning components to tackle this problem. By ensembling all of these models, we developed a final system that outperforms previous models.

Despite these advancements, Code-Mixed ERC remains a challenging task with significant potential for further investigation. Future research directions could include designing robust encoders capable of processing code-mixed dialogues and predicting emotions in an end-to-end manner. Moreover, collecting more data on these code-mixed dialogues is necessary to improve the performance of models. Furthermore, we can explore more complex models that incorporate different information from various modalities to achieve better performance. This work serves as a stepping stone towards more sophisticated emotion recognition systems for code-mixed dialogues.
\bibliography{acl_latex}




\end{document}